\pdfoutput=1

\documentclass[a4paper,twocolumn]{article}

\usepackage{jsaiac}

\usepackage[dvipdfmx]{graphicx}
\usepackage[dvipdfmx]{color}

\usepackage{listings}

\usepackage{amsmath}

\usepackage[export]{adjustbox}

\usepackage{color}

\title{Assessing the Aesthetic Evaluation Capabilities of GPT-4 with Vision: \\
Insights from Group and Individual Assessments}

\address{Yoshia Abe, The University of Tokyo, and\\ ~~~~~~~~~~  y-abe[at]isi.imi.i.u-tokyo.ac.jp}

\author{
Yoshia Abe
\and
Tatsuya Daikoku
\and
Yasuo Kuniyoshi
}

\affiliate{
The University of Tokyo
}

\begin{abstract}
Recently, it has been recognized that large language models demonstrate high performance on various intellectual tasks. 
However, few studies have investigated alignment with humans in behaviors that involve sensibility, such as aesthetic evaluation. 
This study investigates the performance of GPT-4 with Vision, a state-of-the-art language model that can handle image input, on the task of aesthetic evaluation of images. 
We employ two tasks, prediction of the average evaluation values of a group and an individual’s evaluation values. 
We investigate the performance of GPT-4 with Vision by exploring prompts and analyzing prediction behaviors. 
Experimental results reveal GPT-4 with Vision’s superior performance in predicting aesthetic evaluations and the nature of different responses to beauty and ugliness. 
Finally, we discuss developing an AI system for aesthetic evaluation based on scientific knowledge of the human perception of beauty, employing agent technologies that integrate traditional deep learning models with large language models. 
\end{abstract}

\def\BibTeX{{\rm B\kern-.05em{\sc i\kern-.025em b}\kern-.08em%
 T\kern-.1667em\lower.7ex\hbox{E}\kern-.125emX}}
\def\JBibTeX{\leavevmode\lower .6ex\hbox{J}\kern-0.15em\BibTeX}
\def\LaTeXe{\LaTeX\kern.15em2$_{\textstyle\varepsilon}$}

\begin{document}
\maketitle

\section{Introduction}

In recent years, Large Language Models (LLMs) have been shown to be effective for a number of tasks, such as Q\&A tasks and logical reasoning tasks.
While these types of intellectual problem-solving abilities have been widely investigated, there have not been many investigations on performance for tasks requiring sensibility \cite{ye2024flask, openai2023gpt4}.
In this paper, we investigate how closely state-of-the-art LLMs align with humans in one of the sensibility behaviors, aesthetic evaluation.

Humans perform aesthetic evaluations on information accross various modalities. 
In addition to aesthetic evaluation of textual information (e.g. stories and poems), which can be handled by LLMs, 
humans also perceive beauty in images, music, and moral behaviors. 
In this paper, we focus on the beauty of images, a modality where substantial knowledge about aesthetic evaluations has been established.

Recently, GPT-4, one of the most powerful LLMs, has begun to support image input \cite{openai2023gpt4, yang2023dawn}. 
Among various types of Vision Language Models that can handle images and text, in this paper, we adopt GPT-4 with vision (GPT-4V) as the language model under investigation.

The purpose of this paper is to investigate the aesthetic evaluation performance of GPT-4V. 
A comparison of prompt engineering methods and an analysis of the behavior of predictions are also conducted.
The experimental results reveal the nature of the current LLM's aesthetic evaluation performance and behavior, and provide insight for improvement. 
In the last of this paper, we discuss the direction of future developments: we should take scientific knowledge of beauty into account and build an aesthetic evaluation model using agent technology that integrates traditional deep learning models and LLMs.

\section{Related Works}

There has been a lot of research on approximating the aesthetic evaluation of images using AI technologies \cite{deng2017image}. 
Initially, image features were desined directly by humans \cite{ke2006design}, or by machine learning models using generic image features such as the Fisher vector \cite{marchesotti2011assessing}. 

In the past decade or so, deep learning technology has advanced, and research has emerged on approximating the aesthetic evaluation of images using neural networks in a black box manner. 
For example, studies have been reported that use convolutional neural networks to acquire image features through learning \cite{kao2015visual, talebi2018nima}. 
These studies have been contributed to by the fact that large-scale aesthetic evaluation datasets (e.g. AVA \cite{murray2012ava}) were constructed using platform sites on the Internet. These large datasets are beneficial for deep learning, which requires a large amount of training data. 
Most of the above studies set the mean value of the people's aesthetic evaluation as the ground truth for regression or classification. 
This is called the Generic Image Aesthetics Assessment (GIAA) \cite{li2020personality}.

On the other hand, there has been recent progress in adressing the task of approximating the aesthetic evaluation of a particular individual, which is called the Personalized Image Aesthetics Assessment (PIAA) \cite{li2020personality}. 
Several methods have been proposed for constructing PIAA models using deep learning: 
a method where a model that takes an image as an input, receives the individual's attributes as conditioning, and outputs the individual's aesthetic evaluation \cite{yang2022personalized}.  Another method is to firstly train a GIAA model using a dataset of people's aesthetic evaluations and then convert into a PIAA model by fine-tuning with personal data \cite{yang2022personalized}. Another method is to use the GIAA model as a sub-module, make a sub-module that predicts the difference between the output values of the GIAA model and the individual aesthetic evaluations, and build a PIAA model by adding these two sub-modules \cite{ren2017personalized}.

However, the problem with these methods is that the use of the average value of the aesthetic evaluation of the masses neutralizes the differences in individual processes of perception of beauty \cite{zhu2022personalized}.
In the PIAA task, there are the problem of such individual differences in aesthetic evaluation \cite{ren2017personalized} and the problem of the small amount of data for each individual person.
These problems are currently hindering the construction of PIAA models because they are incompatible with deep learning, which requires a large amount of data and consistent labeling. 

One possible solution to such problems might be to use LLMs. 
This is because the perception of beauty is said to be strongly influenced not only by the objective features of the input stimuli, but also by the prior and background knowledge that a person has acquired through past experiences \cite{chatterjee2016neuroscience, reber2004processing}. 
LLMs have extensive knowledge acquired from vast amount of text data, which may be transferable to aesthetic evaluation. 
Therefore, as a first step, this study aims to investigate the current LLM's performance for aesthetic evaluation.

\section{Methods}

Using the OpenAI API of GPT-4V, we investigate how well a LLM performs on aesthetic evaluation of images. 

Two tasks, GIAA and PIAA, are employed. 
GIAA is a task in which GPT-4V predicts the average aesthetic evaluation of an image given by a group of users, as a three-class classification problem.
PIAA is a task in which GPT-4V understands the tendncy of a user's aesthetic evaluation using few-shot prompting, and predicts the user's aesthetic evaluation for a newly presented image based on the understanding. This is also a three-class classification problem. 

PARA \cite{yang2022personalized} exists as a dataset of aesthetic evaluation of images that can be used for both GIAA and PIAA. 
PARA stores aesthetic evaluations by 438 raters for a total of 31220 images. 

In both cases of GIAA and PIAA, basically, the performance indicator is the percentage of correct answers in the three-class classification problem (accuracy). 
This indicator is appropriate because the number of samples in each class (low/middle/high) is the same in our experiments. 
In some cases, we also report the precision, recall, and f1-score for each class as additional performance measures.

\section{Exp. 1: GIAA}

PARA dataset contains images  of 10 semantic patterns: portrait, animal, scene, building, still life, night scene, food, indoor, plant, and others. 
In this experiment, we use six of them (portrait, animal, scene, night scene, food, and plant), which we call "target-semantics". 

PARA dataset stores the average aesthetic score (1-5 floating point number) of the aesthetic evaluation by multiple raters for each image. 
In this experiment, this value is divided into the following three intervals and used as the correct label for the aesthetic evaluation. 
The values greater than 1.0 and up to 2.5 are defined as "low", greater than 2.5 and up to 3.5 are defined as "middle", and greater than 3.5 and up to 5.0 are defined as "high".

Considering the number of patterns to be explored and the GPT-4V's API usage fee, it is not efficient to obtain GPT-4V's aesthetic evaluation for all images in the PARA dataset, so a subset of the entire dataset is used in this experiment. 
Among the images belonging to the 6 patterns of target-semantics, $n$ images are extracted from each of the aesthetic evaluation labels (low/middle/high). 
The resulting $6 * 3 * n = 18n$ image are used as the subset for validation to measure the aesthetic evaluation performance of GPT-4V. 
This is done for 5 random seeds to account for variation.


The prompt for GPT-4V is designed as follows. 
First, the following text is input as a system attribute prompt that presents the premise of the interaction. 

\begin{quotation}
    \texttt{\normalsize{
    You will be asked to provide an aesthetic evaluation of the image that will be presented to you.
    The aesthetic evaluation of the image will be conducted in three levels: 0: low, 1: middle, 2: high.
    First, the image will be presented, followed by a question. Provide your answer after the question is presented.
    When answering, output only in JSON format. Refrain from outputting anything other than JSON format.
    }}
\end{quotation}

Next, after inputting a tokenized image to be evaluated, we input the following text as a user attribute prompt for question about aesthetic evaluation. 
GPT-4V's output is limited to JSON format, so it can be parsed mechanically. 
We call the following prompt the $Q_{\mathrm{Default}}$ prompt to distinguish it from the variations used in Exp. 1-B and 1-C. 

\begin{quotation}
    \texttt{\normalsize{
    Predict the aesthetic evaluation of the image by choosing from three classes (0: low, 1: middle, 2: high) as an integer value (0,1,2), and explain the reason and your confidence level (as a floating-point number ranging from 0 to 1).\\
    \\
    Note:\\
    - Answer in English. \\
    - Answer from the perspective of your position.\\
    - Ensure that your output is strictly in JSON format without any additional characters or formatting symbols such as ``` before and after the JSON object.\\
    - Strictly adhere to the example format provided below:\\
    \{"reason": "\textasciitilde\textasciitilde\textasciitilde", "prediction": 0, "confidence": 0.5\}\\
    }}
\end{quotation}

In making the prompts, we employ various methods of prompt engineering that are often used: clear and specific task instruction, output format settings, example responses, task segmentation and so on. 
In addition, since it is said that GPT-4V currently performs better in English, all the information is written in English.

\subsection{Exp. 1-A: Comparison of Resolution}

GPT-4V can receive tokenized images with text data as input. 
When tokenizing images, two resolution modes, low and high, are available (as of 2024/2). 
When the resolution mode is low, the image is resized to 512x512 and tokenized. 
When the resolution mode is high, the image is reduced to within 2048x2048 with a fixed aspect ratio, then resized to 768 on the short side and tokenized.
In Exp. 1-A, we examine the difference in the prediction performance of the aesthetic evaluation of an image when the resolution mode is set to low or high. 

\begin{figure}[htbp]
  \centering
  \includegraphics[width=0.75\columnwidth, trim=0 0 0 0, clip]{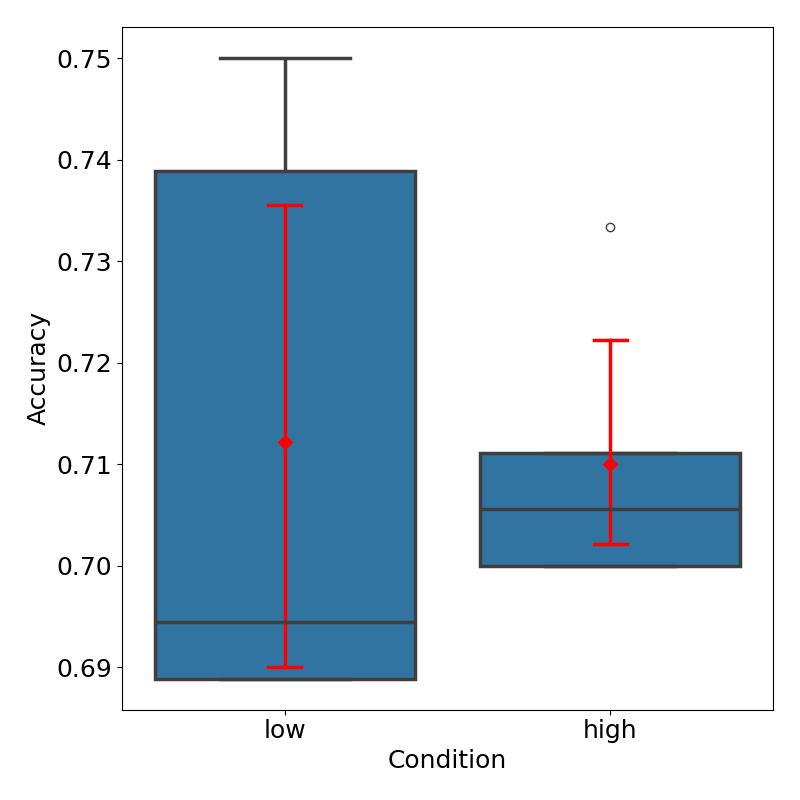}
  \caption[GIAA Detail]{Comparison of accuracy in different resolusion modes}
  \label{fig:exp1_giaa_detail}
\end{figure}

The GIAA performance for the two resolution modes low/high is shown in Fig. \ref{fig:exp1_giaa_detail}. 
This graph shows a box-and-whisker plot of the percentage of correct responses (accuracy) for each of the five seed trials. 
The upper and lower whiskers indicate the range of $1.5$ times the quartile range from the lower and upper quartiles, respectively, and values outside this range are indicated by white circles as outliers. 
The red diamonds indicate the mean values, and the red upper and lower whiskers indicate the 95\% confidence intervals. (The box-and-whisker plot and the red dots are the same for subsequent experiments.) 
A Welch's t-test (significance level $0.05$) was conducted on the sample group of the mean percentage of correct responses (accuracy) in these two conditions, and the p-value was $0.885$, indicating that there was no significant difference. 
Since these results indicate that there is no difference between low and high resolution modes, we decided to use the low resolution mode for all the subsequent experiments.

\subsection{Exp. 1-B: Comparison of Question}

It is well known that when we ask LLMs to perform some reasoning task, it is better to divide the question into subproblems \cite{zhou2023leasttomost} or to make them think in logical steps \cite{wei2023chainofthought}. 
Although it is debatable whether or not logical reasoning is necessary for the image aesthetic evaluation task we are dealing with now, we assume that when a person evaluates an image aesthetically, he/she evaluates the various items first, integrates them, and then makes a final judgment. 
We create two patterns of prompts that follow this assumption. 
In the first pattern $Q_{\mathrm{Elements}}$, GPT-4V is asked to evaluate eight items (composition, shade, color, lightness, darkness, sharpness, depth of field, emotion, and creativity) that can be related to the aesthetic evaluation of an image first, and then answer the final aesthetic evaluation based on these  evaluations. 
In the second pattern $Q_{\mathrm{Techniques}}$, the participants are asked to evaluate the three perspectives known as the basic techniques for taking beautiful pictures \cite{deng2017image} first, and then answer the final aesthetic evaluation based on these evaluations. The three perspectives are: rule of thirds (dividing the image vertically and horizontally into three sections each and placing the subject on the four intersections of the dividing lines to improve balance), low depth of field (showing the subject clearly and blurring the background to make the subject stand out), and opposing colors (combining colors that are opposite to each other on the color wheel to create a strong visual contrast). 

\begin{figure}[htbp]
  \centering
  \includegraphics[width=0.9\columnwidth, trim=0 0 0 0, clip]{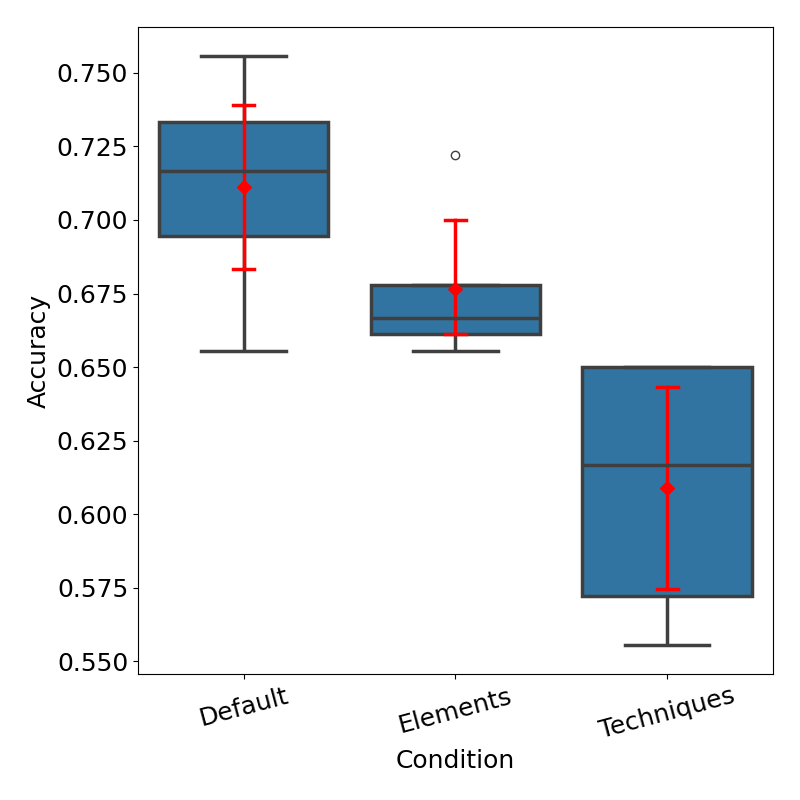}
  \caption[GIAA Question]{Comparison of accuracy in different question cases (GIAA) }
  \label{fig:exp1_giaa_question}
\end{figure}

We compared the three prompt patterns $Q_{\mathrm{Elements}}$, $Q_{\mathrm{Techniques}}$, and $Q_{\mathrm{Default}}$. The results are shown in Fig. \ref{fig:exp1_giaa_question}. 
A Welch's t-test (significance level $0.05$) was conducted on the sample groups of any two conditions taken from the groups of the three conditions, and the results were as follows. 
($Q_{\mathrm{Default}}$, $Q_{\mathrm{Techniques}}$) pair had p-value of $0.004$ and
($Q_{\mathrm{Elements}}$, $Q_{\mathrm{Techniques}}$) pair had p-value $0.022$, indicating a significant difference. 
Although in the ($Q_{\mathrm{Default}}$, $Q_{\mathrm{Elements}}$) pair, the p-value was $0.142$, indicating  no significant difference, the graph shows that the mean and its 95\% confidence interval were higher in the $Q_{\mathrm{Default}}$ condition. 
These results indicate that $Q_{\mathrm{Default}}$ is the best question prompt, so we decided to use $Q_{\mathrm{Default}}$ prompt in the subsequent experiment. 

\subsection{Exp. 1-C: Comparison of Persona}

It is know that it is better to specify a persona when having LLMs perform some inference task \cite{xu2023expertprompting}. 
In Exp. 1-A and 1-B, we investigated the GIAA performance of GPT-4V without specifying a persona, but in Exp. 1-C, we investigate the difference depending on the persona specified. 

Persona information is given at the beginning of the system attribute prompt.
We use 6 patterns of personas: photographer, artist, critic, ordinary person+ (a person interested in photography and painting), ordinary person- (a person not interested in photography or painting), and none (no persona specification). 

\begin{figure}[htbp]
  \centering
  \includegraphics[width=0.9\columnwidth, trim=0 0 0 0, clip]{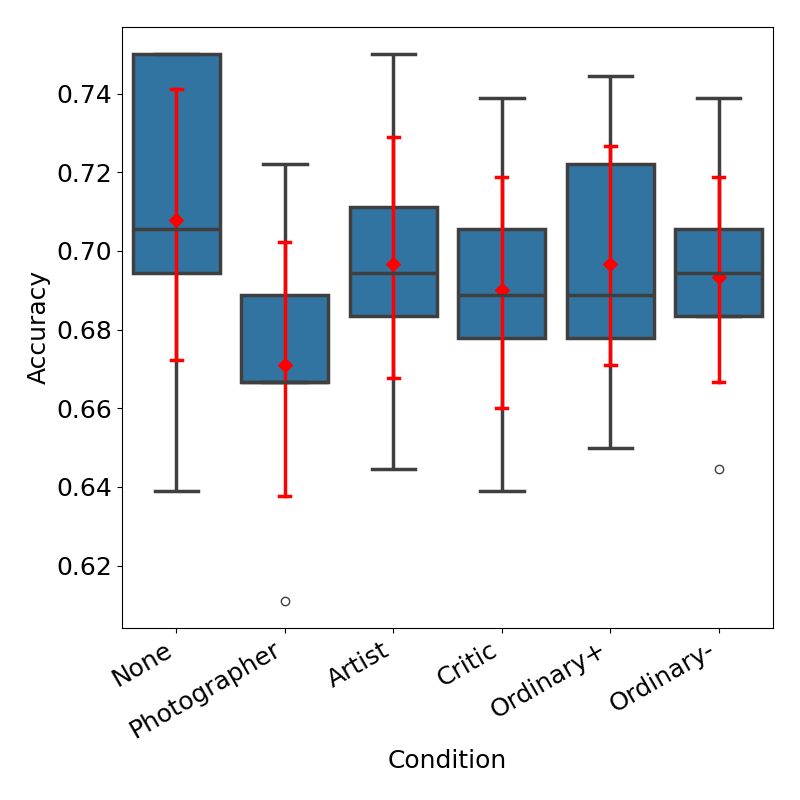}
  \caption[GIAA Persona]{Comparison of accuracy in different persona cases}
  \label{fig:exp1_giaa_persona}
\end{figure}

The result of the comparison of these six personas is shown in Fig. \ref{fig:exp1_giaa_persona}. 
A one-way ANOVA (Analysis of Variance) (significance level $0.05$) was conducted to test for significant differences among the six conditions for the sample group of the mean accuracy. 
The result was p-value $0.788$, indicating no significant difference. 
However, considering the variation in accuracy seen in Fig. \ref{fig:exp1_giaa_persona}, we can see that the case with no persona specification (None) may be relatively the most effective. 
In this case, the prediction performances other than accuracy for the three aesthetic evaluation classes are shown in Table \ref{tab:exp1_giaa_persona=None}. 
All of these are averages over $5$ seeds, rounded to the third decimal place. 
For each indicator, maximum value is shown in bold and the minimum value in underline. 

The precision, recall, and f1-score indices all show that the performance of the other two classes is higher than that of the "middle" aesthetics evaluation class. 
This indicates the obvious phenomenon that the classes with extreme ratings are more predictable than the class with intermediate ratings, regardless of whether they are superior or inferior. 
On the other hand, a different phenomenon is observed for precision and recall. 
Precision, i.e., the percentage of instances predicted as a certain class that actually belong to that class, is higher for "high" than for "low". 
Conversely, the recall, i.e., the proportion of correctly predicted instances of a class among the instances actually belonging to that class, shows that "low" is higher than "high".
These results suggest that the reliability of judgments of beauty is high and ugliness is easy to detect. 
These results indicate the different nature of beauty and ugliness in the GPT-4V's GIAA.

\begin{table}[htbp]
    \centering
    \begin{tabular}{c||c|c|c}
        class & precision & recall & f1-score \\
        \hline
        0: low & 0.771 & \textbf{0.860} & \textbf{0.813} \\
        1: middle & \underline{0.563} & \underline{0.600} & \underline{0.580} \\
        2: high & \textbf{0.814} & 0.663 & 0.730 \\
        \hline
    \end{tabular}
    \caption[GIAA Persona=None]{Comparison of metrics among three classes (GIAA)}
    \label{tab:exp1_giaa_persona=None}
\end{table}

\section{Exp. 2: PIAA}

It is said that when an LLM is asked to perform some inference task, a method that provides several examples of correct answer to the task (called "few-shot examples") improves the performance \cite{brown2020language}. 
This method, few-shot prompting is used in this experiment to make GPT-4V perform the PIAA task.

In this experiment, we measure GPT-4V's PIAA performance using verification subset extracted from PARA dataset, as in Exp. 1. 
Since PARA dataset contains information on what score each user gave to which images, we extract a subset for verification from the set of images evaluated by a user. 
Six semantic patterns in "target-semantics" are used as in Exp. 1.

On the other hand, the correct labels for the aesthetic evaluation are assigned in a different way than in Exp. 1. 
This is because a user's aesthetic evaluation of an image is a discrete value ranging from $1.0$ to $5.0$ in $0.5$ increments. 
The scores of $1.0, 1.5, 2.0$ are "low", $2.5, 3.0, 3.5$ are "middle", and $4.0, 4.5, 5.0$ are "high". 
Among the images in PARA dataset that were evaluated by the user under consideration, for each of the six "target-semantics" patterns, $n$ images are extracted from each of the three aesthetic evaluation classes. 
The resulting $6*3*n = 18n$ images are used as a subset for validation for one user, and the PIAA performance of GPT-4V is measured. 
This is done for $m$ users to account for variation. (In Exp. 2, only one random number seed is used.)

\begin{figure}[htbp]
  \centering
  \includegraphics[width=0.9\columnwidth, trim=0 0 0 50, clip]{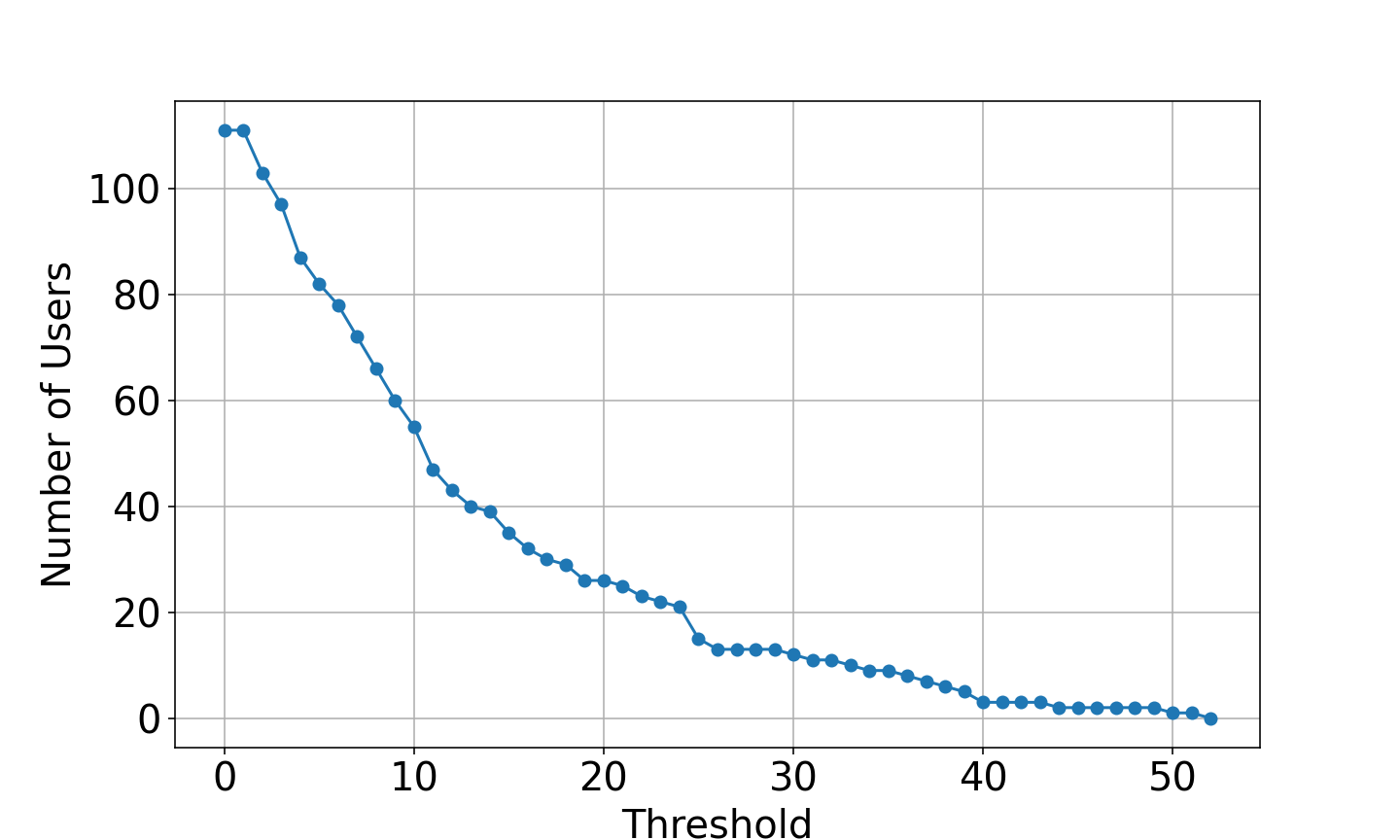}
  \caption[num]{The number of users to be used in Exp. 2}
  \label{fig:exp2_piaa_users}
\end{figure}

Here, the issue is how the number of users for verification "m" is determined. 
In PARA dataset, which consists of the ratings from $438$ users, 
$125$ users rated more than $3000$ images. 
Among these $125$ users, we plot the number of users who have at least threshold $k$ images in each of the 18 patterns (resulting from the product of $6$ semantic patterns and $3$ aesthetic evaluation classes) in Fig. \ref{fig:exp2_piaa_users}.
We consider that it is better to select $k=20$ so that up to $10$ pictures per one pattern in the $18$ patterns can be used as few-shot examples and $10$ pictures can be used for testing. 
So we select the corresponding $m=26$ users as target users for PIAA task (we call them "target-users").

Therefore, for one pattern in the $18$ patterns, the number of images $n $ is $k=20$, the number of candidates of few-shot examples is $10$, the number of few-shot examples that are actually presented $f$ is variable, and the number of test images $t$ is $10$. 
That is, in a subset of $6*3*n = 360$ images for verification for a user, $6*3*t=18t=180$ images are test images, $6*3*10 = 180$ images are candidates of few-shot examples, and $6*3*f=18f$ images are actually used as few-shot examples. 

The presentation of few-shot examples and testing are conducted independently for each semantic patterns. For example, when portraits is used as a semantic pattern, we present $f$ portrait images for each of three aesthetic evaluation classes, as few-shot examples. 
GPT-4V predicts the aesthetic evaluation of newly presented portrait images in the test set based on the understanding of the tendency gained from $3f$ images. 

The prompt is designed as follows. 
First, the following text is input as the system attribute prompt to show few-shot examples.  

\begin{quotation}
    \texttt{\normalsize{
    You will be tasked with understanding a user's "aesthetic evaluation tendencies towards images." 
    A total of \{$3f$\} images will be presented for this purpose, all of which are themed around \{$name ~of ~semantic$\}.
    The aesthetic evaluation of the images will be conducted in three levels (0: low, 1: middle, 2: high).
    }}
\end{quotation}
Then, the tokenized images for few-shot examples and their aesthetic labels made by the target user are input as a system attribute prompt. 

\begin{quotation}
    \texttt{\normalsize{
    \{$Tokenized ~Image$\} \\
    The aesthetic evaluation of the above image is \{$Aesthetic ~Label$\}.
    \#\#\#\#\#\#\#\#\#\#\#\#\#\#\#\#\#\#\#\#\#\#\#\#\#\#\#\#\#\#\#\#\#\#\#\#\#\#\#\#\#\#\#\#\#
    }}
\end{quotation}
Finally, the following text is input as a system attribute prompt. 
\begin{quotation}
    \texttt{\normalsize{
    Based on the examples provided, you will now be asked to predict the user's aesthetic evaluation of the images that will be presented.
    }}
\end{quotation}

Next, after inputting a tokenized image to be evaluated, the following text is input as a user attribute prompt. 
GPT-4V's output is limited to JSON format, so it can be parsed mechanically. 

\begin{quotation}
    \texttt{\normalsize{
    Based on the initial \{$3f$\} example images presented and your interpretation of the user's aesthetic evaluation tendencies, evaluate the aesthetic value of the last image shown.\\
    Predict the aesthetic evaluation of the image by choosing from three classes (0: low, 1: middle, 2: high) as an integer value (0,1,2), and explain the reason and your confidence level (as a floating-point number ranging from 0 to 1).\\
    Note:\\
    - Answer in English.\\
    - Answer from the perspective of the user's position.\\
    - Ensure that your output is strictly in JSON format without any additional characters or formatting symbols such as ``` before and after the JSON object.\\
    - Be careful not to make any mistakes in the JSON format key names.\\
    - Strictly adhere to the example format provided below:\\
    \{"reason": "\textasciitilde\textasciitilde\textasciitilde", "prediction": 0, "confidence": 0.5\}\\
    }}
\end{quotation}

\subsection{Exp. 2-A: Comparison of Question}

As mentioned in Exp. 1-B, it is known that if we want LLMs to perform some reasoning task, it is better to divide the question into subproblems \cite{zhou2023leasttomost} or to think in logical steps \cite{wei2023chainofthought}.
In this PIAA task, we hypothesize that it would be better to ask the GPT-4V to understand the tendency in the few-shot example images and their ratings from the user, and then to ask GPT-4V to predict the aesthetic evaluation of the newly presented test image. 
In Exp. 2-A, we investigate whether the performance on the PIAA task varies depending on the presence or absence of the tendency understanding question. The number of few-shot examples $f$ is set to $3$. 
When asking the tendency understanding, the following prompt (user attribute) is used after presenting few-shot examples.
\begin{quotation}
    \texttt{\normalsize{
    Please describe the user's 'aesthetic evaluation' tendencies towards images, based on the \{$3f$\} images illustrated above.
    }}
\end{quotation}

In this experiment, data for $2$ users out of $26$ target-users are excluded because the images used were sometimes rejected due to GPT-4V's content restriction policy, and the data for the remaining $24$ users are used in the analysis.

\begin{figure}[htbp]
  \centering
  \includegraphics[width=0.75\columnwidth, trim=0 0 0 0, clip]{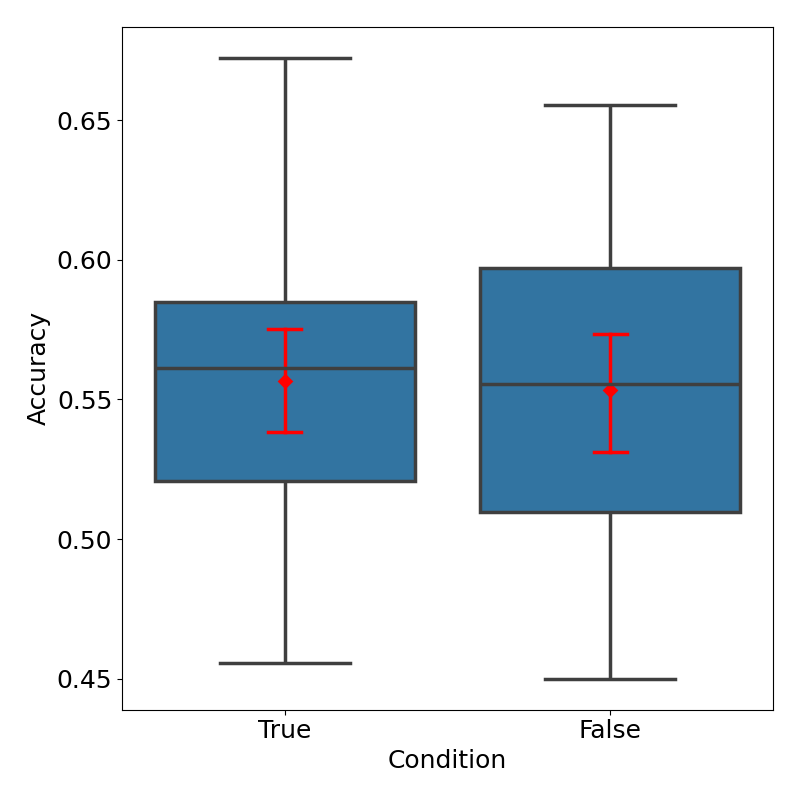}
  \caption[PIAA Question]{Comparison of accuracy in different question cases (PIAA) }
  \label{fig:exp2_piaa_question}
\end{figure}

The difference in PIAA performance with and without the tendency understanding question is shown in Fig. \ref{fig:exp2_piaa_question}. 
"True" and "False" in the figure correspond to the cases with and without the question, respectively. 
A Welch's t-test (significance level $0.05$) was conducted on the sample group of the mean accuracy in these two conditions, and the p-value was $0.822$, indicating no significant difference. 
Since these results indicate that there is no difference in the performance with and without the tendency understanding question, 
we decided to use the condition of "True" (with the question) in the subsequent experiment.

\subsection{Exp. 2-B: Comparison of Few-Shots}

The few-shot example in this PIAA task is important information for understanding the tendency of a target user's aesthetic evaluation. 
We hypothesize that the amount of this information will affect how well GPT-4V can predict the user's aesthetic evaluation. 
In Exp. 2-B, we investigate whether there is a difference in PIAA evaluation performance depending on the number of the few-shot examples $f$ by varying $f=0,3,6,9$.

\begin{figure}[htbp]
  \centering
  \includegraphics[width=0.9\columnwidth, trim=0 0 0 0, clip]{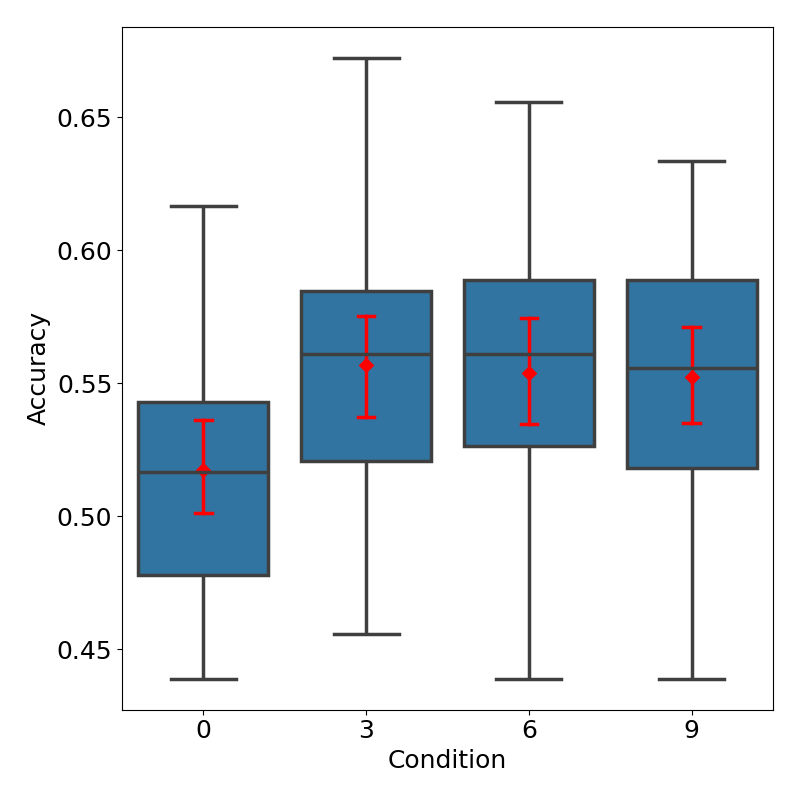}
  \caption[PIAA NFSE]{Comparison of accuracy in different $f$ cases}
  \label{fig:exp2_piaa_nfse}
\end{figure}

In this experiment, in the case of $f=3$, we use the same data obtained in the condition of "True" in Exp. 2-A, so the data used for the analysis is from the $24$ users. 
The difference in PIAA performance due to the difference in the number of few-shot examples is shown in Fig. \ref{fig:exp2_piaa_nfse}. 
A Welch's t-test (significance level $0.05$) was conducted on the two sample groups of the mean accuracy from the sample groups of the four conditions. 
The p-values for the three pairs of $(0,3), (0,6), (0,9)$ were $0.007, 0.014, 0.015$, respectively, indicating significant differences. 
On the other hand, the p-values for the three pairs of $(3,6), (3,9), (6,9)$ were $0.837, 0.765, 0.931$, respectively, indicating no significant differences.

These results indicate that there is a difference between using no examples and using examples with a few shots, but that if examples are used, there may not be a large difference in the number of examples with a few shots. 
However, the number of few-shot examples is only $3, 6, 9$ in this experiment, which is not a broad generalization. 
Perhaps performance could be improved if the number of examples were $50$ or $100$, or if the number of examples were $1$ or $2$. 
There must be a tradeoff between the need for more information to capture individual tendencies and the fact that a larger context length would degrade LLM's performance on inference tasks, so it will be important to find the best parameter $f$. This should be addressed in future works.

Table \ref{tab:exp2_piaa_nfse=3} shows the prediction performance, other than accuracy, for the three aesthetic evaluation classes, when $f=3$.
All of these are averages over $24$ target-users, rounded to the third decimal place. 
For each indicator, maximum value is shown in bold and the minimum value in underline. 
The maximum value for each indicator is shown in class "high" for precision, and in class "low" for recall and f1-score, as in the case of GIAA (Table \ref{tab:exp1_giaa_persona=None}). 
This indicates that the reliability of judgments of beauty is high and ugliness is easy to detect.

On the other hand, the trend of the mimimum value is different from that of the GIAA case. 
In the case of GIAA, the minimum value corresponds to class "middle" for all indicators, while in the case of PIAA, the minimum value of recall was seen in the class "high". 
The fact that the class "high" is the largest in precision but the smallest in recall indicates that it is difficult to determine the tendency of what an individual perceives as beauty, probably because of individual differences, and that if the tendency is understood, the reliability of the prediction is high. 
On the other hand, the high recall of the class "low" is considered to indicate that it is easy to grasp the tendency of ugliness, i.e., there are few individual differences. 
These results of the PIAA experiments also show the different nature of beauty and ugliness in aesthetic evaluation.

\begin{table}[htbp]
    \centering
    \begin{tabular}{c||c|c|c}
        class & precision & recall & f1-score \\
        \hline
        0: low & 0.667 & \textbf{0.622} & \textbf{0.640} \\
        1: middle & \underline{0.418} & 0.585 & \underline{0.487} \\
        2: high & \textbf{0.702} & \underline{0.464} & 0.555 \\
        \hline
    \end{tabular}
    \caption[PIAA NFSE=3]{Comparison of metrics among three classes (PIAA)}
    \label{tab:exp2_piaa_nfse=3}
\end{table}

\section{Discussion}

The purpose of this paper was to investigate the performance of current LLMs in aesthetic evaluation of images. 
There are two different aesthetic evaluation tasks, GIAA and PIAA, which were tested in Exp. 1 and 2, respectivey. 
We used the percentage of correct response (accuracy) to a three-classes classification task as an index to evaluate performance. 
At the same time, since the performance of LLMs is known to vary widely across prompts, we examine the image resolution mode, the question items, and the persona patterns in Exp. 1, and the influence of question of understanding target user's aesthetic evaluation tendency and the number of few-shot examples in Exp. 2.

The results of Exp. 1 show that there is no difference in performance between different image resolution modes in GPT-4V's API. 
The results also show that performance is higher when no questions were used and no persona was specified. 
This suggests that because GIAA is a task to predict the average aesthetic evaluation of the group of users, the addition of question items or personas may lead to a limitation or misdirection of perspectives and thoughts. 
The performance on the GIAA task in the best condition, in other words, in the case where no question is used and no persona is specified in Exp. 1-C, was $0.708$ (rounded to the third decimal places) in accuracy. 
The fact that, although GPT-4V has not been specially trained for the task of aesthetic evalution of images, it achieved a correct response rate of $0.708$, which is much higher than the chance level $0.333$ for the three-classes classification task, is surprising.

The results of Exp. 2 show that there is no difference in performance depending on whether or not the question is about understanding individual tendencies, and that as long as the number of few-shot examples is not zero, similar performance levels are demonstrated.
The performance on the PIAA task in the best condition, in other words, in the case where GPT-4V is asked a question of tendency understanding and the number of few-shot examples $f=3$ in Exp. 2-B, was $0.557$ (rounded to the third decimal places) in accuracy. 
Compared to the GIAA task where the average aesthetic evaluation is predicted, it is natural that the accuracy is lower for the PIAA task where individual differences must be considered.
Nevertheless, it is again surprising that GPT-4V achieved a correct response rate of $0.557$ for the three-classes classification task, which is much higher than the chance level of $0.333$.

The analysis of indices other than accuracy revealed differences in the nature of beauty and ugliness in the aesthetic evaluation of GPT-4V for both GIAA and PIAA. 
The results show that the prediction performance is higher when evaluations are biased towards either beauty or ugliness, and that the reliability of judging beauty is higher, while ugliness is easier to detect. 
These results suggest that different approaches may be needed for beauty and ugliness in the same aesthetic evaluation task.

The limitations of the experiments in this paper are as follows. 
First, the experiments in this paper explored what kind of prompts are best for GIAA and PIAA, but the search range was limited to question items, personas, and the number of few-shot examples. 
A lot of new methods have been proposed for designing prompts, so it is necessary to incorporate them. 
In addition, we only used GPT-4V as a validation target, so comparisons with other language models and deep learning-based models are needed. 
Furthermore, more detailed understanding of the prediction tendencies, other than the comparisons of prediction performance among classes, is required: the analysis of output texts, image features, and user personality traits. 
If we can gain insight into human perception of beauty through the analysis of the behavior of LLMs, it would be an excellent scientific contribution.

Experiments in this paper show that the current LLM, especially GPT-4V, perform well to some extent on GIAA tasks with accuracy more than $70$\%, even though it has not been trained specifically for the aesthetic evaluation task. 
This suggests the possibility that, in creating a superior aesthetic evaluator, there is an effective strategy not only in training a black-box neural network from scratch but also in using LLMs that already may possess some structure of aesthetic evaluation as a foundation. 
The method of appropriately integrating conventional deep learning models and LLMs to improve the behavior of an AI system is known as agent technology \cite{wang2023survey} in recent years. In the context of creating AI models for aesthetic evaluation, research should be accelerated in this direction in the future.

This also points to the importance of model-based development that assumes some structure in the aesthetic evaluation process, rather than model-free construction that does not assume any structure to the evaluation process. 
An LLM trained on the vast amount of text data on the Internet may already have some structure embedded in it. 
It can be compared to a fabric woven from countless intricately intertwined threads.
Within this, we have to carefully find and unravel a thread that closely approximates the aesthetic evaluation process. 
This would contribute to creating a better aesthetic evaluator, especially for the PIAA task, because an individual's perception of beauty is influenced by knowledge or contexts \cite{chatterjee2016neuroscience}.
For this purpose, a deeper understanding of what factors shape an individual's aesthetic evaluation is also necessary. 
In this regard, AI systems for aesthetic evaluation should be developed based on knowledge of other scientific fields such as the humanities related to beauty and neuroaesthetics.

\section{Conclusion}

The experiments in this paper reveal the performance of GPT-4V on aesthetic evaluation tasks, providing insights for enhancing its effectiveness and highlighting areas for further investigation. 
Future research directions will be to construct AI systems that capture human aesthetic evaluation processes using agent technologies that integrate conventional deep learning models and LLMs, based on scientific knowledge about perception of beauty. \\
\textbf{ACKNOWLEDGMENT} \\
This research was supported by Next Generation AI Research Center of The University of Tokyo and JST SPRING GX project (Grant Number JPMJSP2108).



\end{document}